\documentclass[letterpaper]{article} % DO NOT CHANGE THIS
\usepackage{aaai2026}  % DO NOT CHANGE THIS
\usepackage{times}  % DO NOT CHANGE THIS
\usepackage{amssymb}
\usepackage{adjustbox}  % for alignment
\usepackage{pifont}
\usepackage{lineno}
\usepackage{xcolor}
\usepackage{enumitem}
\usepackage{courier}  % DO NOT CHANGE THIS
\usepackage{booktabs}

\usepackage[hyphens]{url}  % DO NOT CHANGE THIS

\usepackage{helvet}  % DO NOT CHANGE THIS
\usepackage{amssymb}
\usepackage{amsmath}

\usepackage{subfigure}
\usepackage{multirow}
\usepackage[capitalise]{cleveref}
\usepackage{float} % For [H] option

\usepackage{graphicx} % DO NOT CHANGE THIS
\usepackage{natbib}  % DO NOT CHANGE THIS AND DO NOT ADD ANY OPTIONS TO IT
\usepackage{caption} % DO NOT CHANGE THIS AND DO NOT ADD ANY OPTIONS TO IT
\usepackage{algorithm}
\usepackage{algorithmic}

\usepackage{newfloat}
\usepackage{listings}
\DeclareCaptionStyle{ruled}{labelfont=normalfont,labelsep=colon,strut=off} % DO NOT CHANGE THIS
\floatstyle{ruled}
\newfloat{listing}{tb}{lst}{}
\floatname{listing}{Listing}
\title{MoEGCL: Mixture of Ego-Graphs Contrastive Representation Learning \\ for Multi-View Clustering}
\author{
    Jian Zhu\textsuperscript{\rm 1}$^{\dagger}$, 
    Xin Zou\textsuperscript{\rm 2}$^{\dagger}$, 
    Jun Sun\textsuperscript{\rm 1}$^{*}$, 
    Cheng Luo\textsuperscript{\rm 1}$^{*}$, 
    Lei Liu\textsuperscript{\rm 3}, 
    Lingfang Zeng\textsuperscript{\rm 1}, \\
    Ning Zhang\textsuperscript{\rm 1}$^{*}$, 
    Bian Wu\textsuperscript{\rm 1}, 
    Chang Tang\textsuperscript{\rm 4}, 
    Lirong Dai\textsuperscript{\rm 3}
}
\affiliations{
    \textsuperscript{\rm 1} Zhejiang Lab, Hangzhou, China\\
    \textsuperscript{\rm 2} The Hong Kong University of Science and Technology, Guangzhou, China\\
    \textsuperscript{\rm 3} University of Science and Technology of China, Hefei, China\\
    \textsuperscript{\rm 4} Huazhong University of Science and Technology, Wuhan, China\\
    Email: qijian.zhu@outlook.com\\
    $\dagger$ Contributed equally to this work,
    $^{*}$Corresponding Author
    
}

\usepackage{bibentry}
\begin{document}

\maketitle

\begin{abstract}
In recent years, the advancement of Graph Neural Networks (GNNs) has significantly propelled progress in Multi-View Clustering (MVC). However, existing methods face the problem of coarse-grained graph fusion. Specifically, current approaches typically generate a separate graph structure for each view and then perform weighted fusion of graph structures at the view level, which is a relatively rough strategy. To address this limitation, we present a novel Mixture of Ego-Graphs Contrastive Representation Learning (MoEGCL). It mainly consists of two modules. In particular, we propose an innovative Mixture of Ego-Graphs Fusion (MoEGF), which constructs ego graphs and utilizes a Mixture-of-Experts network to implement fine-grained fusion of ego graphs at the sample level, rather than the conventional view-level fusion. Additionally, we present the Ego Graph Contrastive Learning (EGCL) module to align the fused representation with the view-specific representation. The EGCL module enhances the representation similarity of samples from the same cluster, not merely from the same sample, further boosting fine-grained graph representation. Extensive experiments demonstrate that MoEGCL achieves state-of-the-art results in deep multi-view clustering tasks. The source code is publicly available at \url{https://github.com/HackerHyper/MoEGCL}.
\end{abstract}


\section{Introduction}
	\label{sec.introduction}
With the rapid advancement of digitalization, data is increasingly being collected from multiple heterogeneous sources \cite{zou2024look,zoudon,zhu:82}. For example, autonomous vehicles rely on multi-camera systems to capture diverse visual perspectives for decision-making. In structural biology, proteins exhibit complex quaternary structures that can be characterized through different analytical techniques. Modern medical diagnostics integrate multimodal data from various clinical tests to enhance accuracy. The term ``multi-view data" is used to describe an item from multiple data sources. Multi-View Clustering (MVC) ~\cite{Tang:65,zou2023inclusivity,xiao2024dual} aims to integrate heterogeneous data from multiple views to discover meaningful cluster structures in an unsupervised manner, playing a pivotal role in modern data mining applications. These methods utilize a view-specific encoder network to generate a powerful representation, which is fused from various views for clustering tasks. However, there exists the issue of heterogeneity among data from different views. Many alignment methods have been proposed to solve this issue. For example, some methods use KL divergence to align the label or representation distributions among different views ~\cite{hershey2007approximating,liusparsemvc,zou2023dpnet}. Additionally, MVC employs contrastive learning (CL) to align representations across different views.

\begin{figure}
  \centering
  \vspace{2em}
  \includegraphics[width=8.4cm]{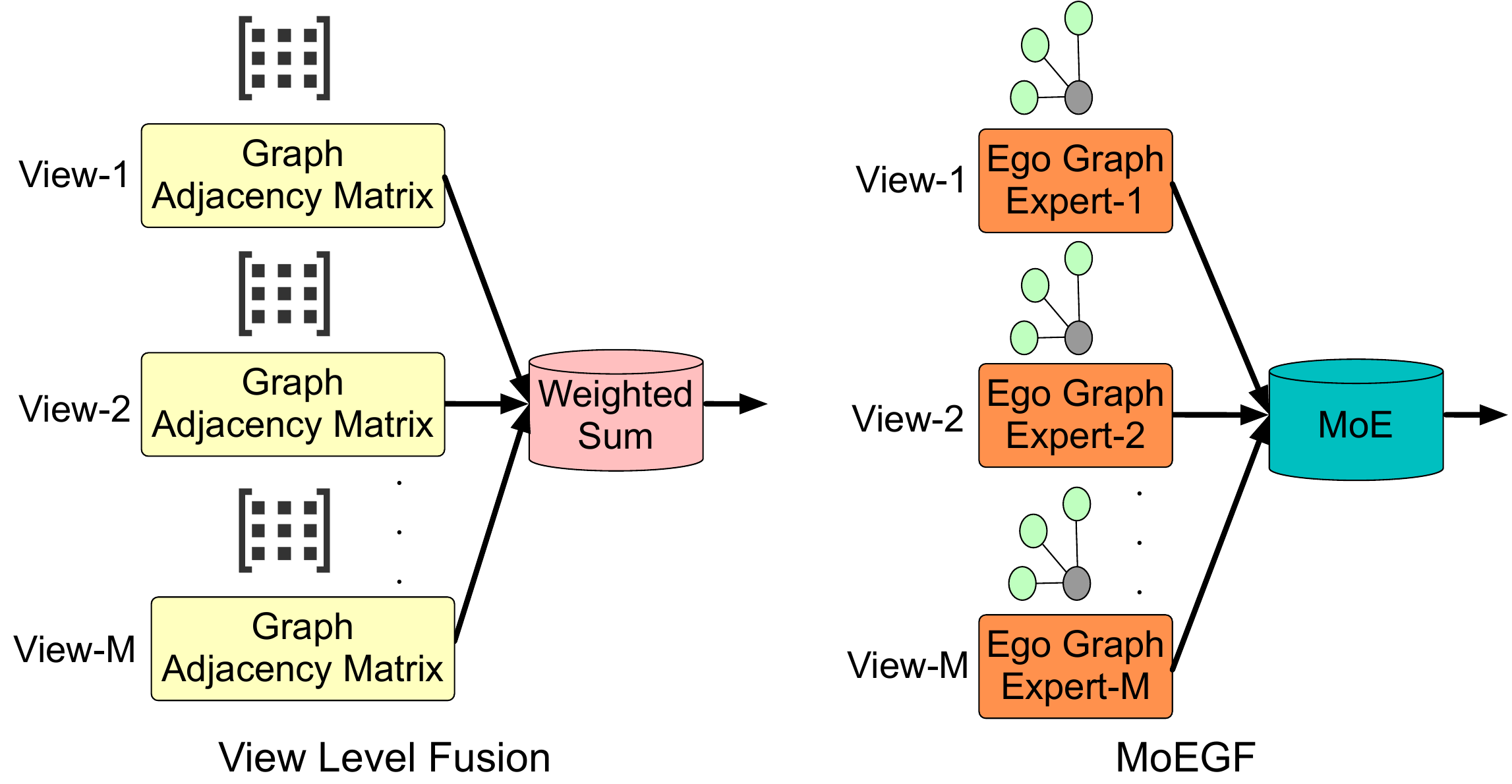}
  \caption{Mixture of Ego-Graphs Fusion. Firstly, an ego graph for each sample in each view is constructed, and input each ego graph as an expert into the Mixture-of-Experts (MoE) architecture and calculate the importance of each expert. Then, the importance coefficient with the experts are merged to form a fused ego graph. Lastly, aggregate the ego graphs of all samples to generate a global connected graph.}
  \label{fig:01}
\end{figure}

In recent years, Graph Neural Networks (GNNs) have emerged as a powerful paradigm for extracting and processing graph-structured information, achieving remarkable success across various domains. Although GNNs have significantly improved the MVC, the problem of coarse-grained graph fusion still exists. For instance, existing methods like DFMVC \cite{ren:75} and CAGL \cite{wang:76} first construct an independent graph for each view and then perform weighted fusion at the view level. However, this approach assigns a fixed fusion coefficient to the graph structure of each view, ignoring the differences between samples. This fusion approach operates at a coarse granularity and fails to capture fine-grained structural relationships, which ultimately limits the performance in deep MVC tasks.

To address the problem of coarse-grained graph fusion, we propose a novel Mixture of Ego-Graphs Contrastive Representation Learning (MoEGCL). It consists of two modules, which are Mixture of Ego-Graphs Fusion (MoEGF) and Ego Graph Contrastive Learning (EGCL). As illustrated in Figure \ref{fig:01}, MoEGF is a new paradigm in deep multi-view clustering. Our proposed paradigm creates ego graphs and uses a Mixture-of-Experts (MoE) architecture \cite{jacobs:55} for the fusion of ego graphs at the sample level. The MoE architecture empowers MoEGF to dynamically integrate information from multiple domain-specific experts. We use the ego graph of each view as an expert in MoE architecture. Different ego graphs can denote the connectivity relationship of a sample in different views. Therefore, the MoEGF module adaptively calculates the confidence of the ego graph, rather than using a fixed weight to measure the graph of each view. This enables better utilization of refined information, achieving fine-grained graph fusion. In addition, we propose the EGCL module to enhance the representation similarity of the samples in the same cluster, rather than the same sample. The EGCL module overcomes the limitations of contrastive learning in current multi-view clustering. It obtains better semantic structure information, thereby improving the fine-grained graph representation.

The main contributions can be summarized as follows:
\begin{itemize}[leftmargin=9.2pt]
\item We present the MoEGF module in the context of multi-view learning. It implements fine-grained graph fusion via a Mixture-of-Experts network at the sample level. 
% To the best of our knowledge, we are the first to propose the fusion of ego graphs.
\item Different with previous approaches that consider different views on a sample to be positive points, we propose the EGCL module to improve the representation similarity of samples within the cluster, which enables fine-grained graph representation.
\item Extensive experiments demonstrate that the MoEGCL module establishes new state-of-the-art performance in deep multi-view clustering, outperforming existing methods by significant margins on the six public datasets.
\end{itemize}

\section{Related Works}
\subsection{Mult-view Clustering}
There are mainly two kinds of methods in Multi-View Clustering (MVC) \cite{Chem2022review,zhu2025generative}: Shallow Multi-View Clustering methods and Deep Multi-View Clustering methods.
Shallow MVC methods \cite{Chem2022review} are classified into two categories: Graph Multi-View Clustering and Subspace Multi-View  Clustering. Graph Multi-View Clustering methods~\cite{fang2022structure,pan2021multi} generally follow these steps: construct view-specific graphs, followed by the creation of a fusion graph from several view-specific graphs using different regularization terms, and finally the generation of the clustering results using spectral clustering, graph-cut approaches, or other algorithms. Subspace Multi-View Clustering methods \cite{Zhang:64, Huang:67} create a common subspace self-representation vector from every view, and then separate the samples into different subspaces using different regularization terms on the consensus self-representation matrix.

Many recent works have focused on Deep MVC, motivated by encouraging developments in deep learning \cite{zhu:81,zhu:83,zhu2025trusted,xu2025hstrans}. Specifically, these methods encode the non-linear feature using a deep neural network. The adversarial methodology is used in Deep MVC methods \cite{li2021multi,wang2022adversarial} to coordinate patterns of hidden representations from different views and learn the latent representation. The attention mechanism is used by \citet{zhou2020end,zhu2025dynamic} to allocate each view a weight value. To obtain a consensus representation, they then use the weighted total of all view-wise presentations. The consensus representation is produced by Wang et al. \cite{wang2022adversarial} using a weighted aggregation and $l_{1,2}$-norm restriction. Contrastive Learning (CL) may align representations from different perspectives at the sample level, which facilitates label distribution alignment. These CL methods~\cite{trosten2021reconsidering, xu2022multi} have outperformed the previous distribution alignment strategies in Multi-View Clustering, yielding state-of-the-art results on numerous public datasets.

\subsection{Mixture of Experts}
Jacobs et al. \cite{jacobs:55} present the first Mixture-of-Experts (MoE) structure. It is viewed as a tree-structured framework that employs the divide-and-conquer tactic. Sparsely-gated MoE \cite{Noam:77} is the first method to show significant gains in model capacity, training time, or model quality when gating is used. MH-MoE \cite{wu:80} divides each input token into many smaller tokens, which are then assigned to and processed by a variety of experts concurrently before being smoothly reintegrated into the original token form.
We use a Mixture-of-Experts network to implement fine-grained graph fusion at the sample level. To the best of our knowledge, we are the first to present the multi-view fusion framework of ego graphs.

\section{The Proposed Methodology}
\label{section:proposed_method}
\begin{figure*}
  \centering
  \includegraphics[width=17.5cm]{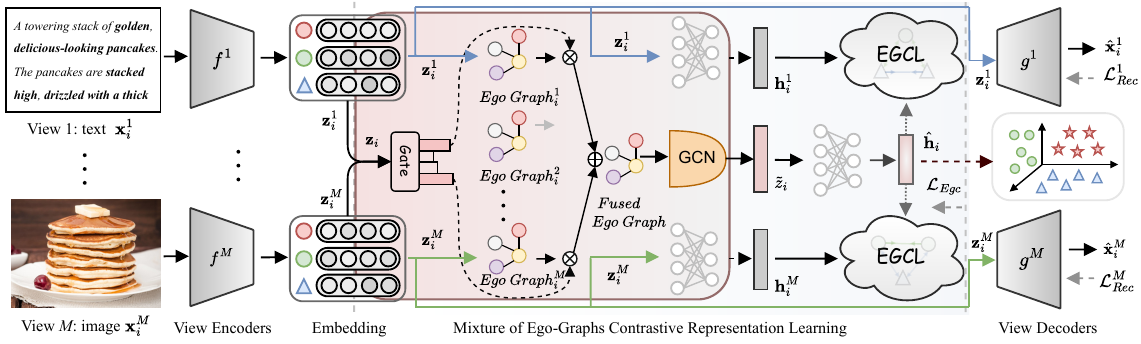}
  \caption{The architecture diagram of MoEGCL. The MoEGCL method contains two modules: MoEGF and EGCL. MoEGF first constructs ego graphs and then aggregates ego graphs of all views into a fused ego graph by a Mixture-of-Experts network. In addition, the EGCL module is proposed to improve the representation similarity of samples in the same cluster, as opposed to only concentrating on the consistency at the sample level. It significantly enhances fine-grained graph representation.}
  \label{fig:02}
\end{figure*}

This paper proposes an innovative Mixture of Ego-Graphs Contrastive Representation Learning (MoEGCL), which aims to overcome the limitation of coarse-grained graph fusion in deep multi-view clustering. As illustrated in Figure \ref{fig:02}, MoEGCL comprises two key components: 1) Mixture of Ego-Graphs Fusion (MoEGF) and 2) Ego Graph Contrastive Learning (EGCL). The multi-view data, which consists of $N$ samples with $M$ views, is denoted as $\{\mathbf{X}^{m}=\{{x}_{1}^{m};...;{x}_{N}^{m}\}\in \mathbb{R}^{N \times D_{m}}\}_{m=1}^M$, where $D_m$ represents the feature dimensionality for the $m$-th view.

\subsection{Autoencoder Network}
We create the representation of each view by using the Autoencoder architecture \cite{song2018self,zou2023hierarchical,zou2024dai}. It contains two parts: an encoder and a decoder. For the $m$-th view, $f^{m}$ represents the encoder. The encoder module generates the representation of each view as follows:
\begin{equation}
\label{eq:encoder}
\begin{aligned}
{z}_{i}^{m}=f^{m}\left({x}_{i}^{m}\right), {z}_{i}^{m} \in \mathbb{R}^{d_{\psi}},
\end{aligned}
\end{equation} 
where ${z}_{i}^{m}$ denotes the representation of the $i$-th instance in the $m$-th view. $d_{\psi}$ represents the feature size of the ${z}_{i}^{m}$.

The $x^m_i$ is restored by the decoder module utilizing the feature ${z}_{i}^{m}$. Let $g^{m}$ represents the decoder module in the $m$-th view. The recovered sample ${\hat{x}}_{i}^{m}$ is produced by decoding ${z}_{i}^{m}$ in the decoder module:
\begin{equation}
\label{eq:decoder}
\begin{aligned}
\hat{{x}}_{i}^{m}=g^{m}\left({z}_{i}^{m}\right).
\end{aligned}
\end{equation}
The reconstruction loss is calculated as follows:
\begin{equation}
\label{eq:reconstruction loss}
\begin{aligned}
\mathcal{L}_{\mathrm{Rec}}=&\sum_{m=1}^{M}\left\|{X}^{m}-\hat{{X}}^{m}\right\|_{F}^{2}\\
=&\sum_{m=1}^{M}\sum_{i=1}^{N}\left\|{x}_i^{m}-g^{m}\left({z}_{i}^{m}\right)\right\|_{2}^{2},
\end{aligned}
\end{equation}
where $||\cdot||_F$ denotes the Frobenius norm.
%$\mathcal{L}_{Rec}$ denotes the reconstruction loss.

\subsection{Mixture of Ego-Graphs Fusion (MoEGF)}\label{sec:ssm}
We present the Mixture of Ego-Graphs Fusion (MoEGF) to achieve fine-grained graph fusion. It consists of three components: Ego Graph Networks, Mixture-of-Experts Network, and Graph Convolutional Network.

\subsubsection{Ego Graph Networks}
First, we create Ego Graph Networks at all views. In particular, we construct the graph $G^m=\left\{Z^m, S^m\right\}$ for the $m$-th view, where $Z^m=\left[z_1^m; z_2^m; \ldots ; z_N^m\right]$ represents the node features of graph $G^m$. $S^m$ is the adjacency matrix. In the $m$-th view, the edge between samples $i$ and $j$ is represented by $S_{ij}^m$. If $S_{ij}^m=1$, samples $i$ and $j$ are connected; otherwise, the two samples are unconnected. We create the adjacency matrix $S^m$ using the KNN graph \cite{peterson:73} as follows:
\begin{equation}
S_{i j}^m= \begin{cases}1 & j \in K_i^m; \\ 0 & \text { otherwise.}\end{cases}
\end{equation}
In the $m$-th view, $K_i^m$ is the collection of $k$ nearest neighbors of $z_i^m$ based on the Euclidean distance.
\begin{equation}
\label{moematrix}
S^m=(V^m_1; V^m_2; \ldots; V^m_N)^{T},
\end{equation}
where $V^m_i$ represents the connection relationship of the ego graph, which is from the $m$-th view of the $i$-th sample. Specifically, it is implemented using the row vectors of the adjacency matrix, also known as the adjacency vector.

\subsubsection{Mixture-of-Experts Network}
We use the Mixture-of-Experts Networks to integrate ego graphs. It consists of a gating network and an expert network. We concatenate the vectors of $M$ views as input to the gating network.
\begin{equation}
z_i=cat(z^1_i; z^2_i; \ldots; z^M_i), z_i \in \mathbb{R}^{Md_{\psi}}.
\end{equation}
We calculate the gating coefficient for each expert with respected to input $z_i$. It is implemented through a multi-layer perceptron neural network $mlp^1$, followed by classification using a softmax activation function. The number of output categories by softmax is $M$. The specific calculation is as follows:
\begin{equation}
\mathcal{C}_i=\operatorname{softmax}(mlp^{1}(z_i))_i.
\end{equation}
Let $V^m_i$ be the $m$-th expert network. Combining the gate network and expert network, a weighted sum is performed to implement the fusion of ego graphs at the sample level:
\begin{equation}
\label{moefused}
V_i=\sum_{m=1}^{M} \mathcal{C}_i^m V^m_i,
\end{equation}
$V_i$ denotes the fused adjacency vector of the $i$-th sample. Aggregating all adjacency vectors can generate an adjacency matrix,
\begin{equation}
\label{moematrix}
S=(V_1; V_2; \ldots; V_N)^{T},
\end{equation}
where $S$ denotes the fused adjacency matrix.

\subsubsection{Graph Convolutional Network}
To improve the clustering results using both global features and structural data, we additionally use the Graph Convolutional Networks (GCN) \cite{Kipf:74,zhang2019graph} to generate boosted graph representations. We adopt a two-layer GCN module to obtain the topological structure information. The graph representation $\tilde{Z}_i$ is computed as follows: 
\begin{equation}
\tilde{Z}=\left(\tilde{D}^{-\frac{1}{2}} \tilde{S} \tilde{D}^{-\frac{1}{2}}\right)\left[\left(\tilde{D}^{-\frac{1}{2}} \tilde{S} \tilde{D}^{-\frac{1}{2}}\right) ZW^{0}\right]W^{1},
\end{equation}
where $\tilde{S}=I_N+S$. $I_N$ is the identity matrix. $\tilde{D}$ is a diagonal matrix, $\tilde{D}_{i i}=\sum_j \tilde{S}_{i j}$. $W^0$ and $W^1$ are the training parameters. $Z=\{z_1; z_2; \ldots;  z_N\}$ is an input feature, and $\tilde{Z}=\{\tilde{z}_1; \tilde{z}_2; \ldots; \tilde{z}_N\}$ is an output feature.

\subsection{Ego Graph Contrastive Learning (EGCL)}
We propose the Ego Graph Contrastive Learning (EGCL) module to improve fine-grained graph representation in deep multi-view clustering tasks. EGCL needs to unify the dimensions of each view feature and the fused feature. The specific calculation is as follows:
\begin{equation}
{\hat{h}}_{i}=mlp^{2}(\tilde{z}_i), {\hat{h}}_{i} \in \mathbb{R}^{d_{\phi}},
\end{equation}
where the dimension of $\tilde{z}_i$ is decreased via the $mlp^2$ operator. $d_{\phi}$ denotes the feature dimension of the ${\hat{h}}_{i}$. Similarly, we minimize dimension on each view feature $z^{m}_i$ using the $mlp^{3,m}$ operator,
\begin{equation}
{{h}^{m}_{i}}=mlp^{3,m}(z^{m}_i), {{h}^{m}_{i}} \in \mathbb{R}^{d_{\phi}}.
\end{equation}
For the ${h}^{m}_{i}$, $d_{\phi}$ is the feature size. The cosine function is used to compute the similarity between common embedding ${\hat{h}_i}$ and view-specific embedding ${h}^{m}_i$:
\begin{equation}
\label{eq:sim}
\begin{aligned}
C\left({\hat{h}}_{i}, {h}_i^{m}\right)= \cos ({\hat{h}}_{i}, {h}_i^{m}).
\end{aligned}
\end{equation}
The loss function of the EGCL module is calculated as follows:
\begin{equation}
\label{eq:lc}
\begin{aligned}
\mathcal{L}_{\mathrm{Egc}}=-\frac{1}{2 N} \sum_{i=1}^{N} \sum_{m=1}^{M} \log \frac{e^{\operatorname{C}\left({\hat{h}}_{i}, {h}_{i}^{m}\right) / \tau}}{\sum_{j=1}^{N} e^{(1-{S}_{ij})\operatorname{C}\left({\hat{h}}_{i}, {h}_{j}^{m}\right) / \tau}},
\end{aligned}
\end{equation}
where $\tau$ denotes the temperature coefficient. Based on Eq. (\ref{moematrix}), we obtain ${S}_{ij}$. We incorporate ${S}_{ij}$ into Eq. (\ref{eq:lc}) to improve the similarity of the samples in the cluster. When $S_{ij}$ is low, the two samples are not from the same cluster. At this point, the $(1-{S}_{ij})\operatorname{C}({\hat{h}}_{i}, {h}_{j}^{m})$ plays a significant role. When $S_{ij}$ is relatively large, it indicates the two samples belong to the same cluster. At this time, the effect of $(1-{S}_{ij})\operatorname{C}({\hat{h}}_{i}, {h}_{j}^{m})$ is small.

The overall loss function is defined as follows:
\begin{equation}
\label{zong}
{\cal L} = {{\cal L}_{\rm{Rec}}} + \lambda {{\cal L}_{\rm{Egc}}},\\
\end{equation}
where $\lambda$ balances the contribution of the two loss terms. Here, ${{\cal L}_{\rm{Rec}}}$ denotes the reconstruction loss, while ${{\cal L}_{\rm{Egc}}}$ represents the ego graph contrastive loss. Our training pipeline consists of two distinct phases: pre-training phase and fine-tuning phase. ${{\cal L}_{\rm{Rec}}}$ is the loss function we use during the pre-training phase, while ${\cal L}$ is the loss function during the fine-tuning phase.

%Previously, contrastive learning in deep multi-view clustering used different view data of an instance as positive samples, while ignoring the similarity of instances within the same cluster. This leads to the optimization of multi-view fusion in the wrong direction. We propose the EGCL module, which breaks through the limitations of contrastive learning at the sample level. It reduces conflicts in representations within the same cluster and enhances similarity within the same cluster, thereby improving fine-grained graph representation.

\subsection{Clustering Module}
We utilize the k-means method as the clustering function \cite{mackay2003information,Yan_2023_CVPR}. Specifically, the common embedding ${\bf{H}}$ has the following factorization:
\begin{equation}
\label{clustering1}
\begin{array}{l}
\begin{array}{*{20}{c}}
{\mathop {\min }\limits_{{\bf{U,V}}} }&{{{\left\| {{\bf{H}} - {\bf{UV}}} \right\|}^2}}
\end{array},  {\bf{H}}=\{\hat{h}_1,...,\hat{h}_N\},\\
s.t.{\bf{U1}} = {\bf{1}},{\bf{U}} \ge {\bf{0}},
\end{array}
\end{equation}
where ${\bf{U}} \in \mathbb{R}^{N \times k} $ denotes the matrix of cluster indicator. We use ${\bf{V}} \in \mathbb{R}^{k\times d^{\phi}}$ as the matrix of clustering center.

\section{Experiments}
In this section, we conduct extensive experiments to evaluate MoEGCL on six benchmark datasets for deep multi-view clustering. To demonstrate its superiority, we compare against eight state-of-the-art baselines across multiple metrics. Furthermore, we perform convergence analysis, visualization, and hyperparameter optimization.

% \subsection{Experimental Settings}
\subsection{Benchmark Datasets}
As illustrated in Table \ref{tab:Datasets}, we utilize six benchmark datasets with various scales for analysis and comparison. These six datasets are Caltech5V, WebKB, LGG, MNIST, RBGD, and LandUse. These six datasets are publicly available and commonly used to evaluate the performance of Multi-View Clustering.

\begin{table}[ht]
\centering
\renewcommand\tabcolsep{2.5pt}
\setlength{\abovecaptionskip}{0.1cm}  %段前
\caption{Summary of the six public datasets.}\small 
\begin{tabular}{lcccc} 
\toprule
{Datasets}    & {Samples} & {Views} & {Clusters} & {\#-view Dimensions}   \\ 
\midrule
Caltech5V  & 1400   & 5     & 7    & [40, 254, 1984, 512, 928]   \\
WebKB  & 1051   & 2     & 2  &  [2949, 334]   \\
LGG  & 267     & 4     & 3   &  [2000, 2000, 333, 209]    \\
MNIST  & 60000     & 3     & 10   &  [342, 1024, 64]    \\
RGBD  & 1449     & 2     & 13   &  [2048, 300]    \\
LandUse & 2100     & 3     & 21   &  [20, 59, 59]    \\
\bottomrule
\end{tabular}
\label{tab:Datasets}
\end{table}

\begin{table*}[t]
\renewcommand\arraystretch{1}
\small
\setlength{\tabcolsep}{9pt}
\setlength{\abovecaptionskip}{0.1cm}  %段前
\centering
\caption{Clustering result comparison on the MNIST, LGG, and WebKB datasets. The best results in bold, second in underline.}
\resizebox{\textwidth}{!}{\begin{tabular}{l|ccc|ccc|ccc} 
\toprule
\multirow{2}{*}{\textbf{Methods}}        & \multicolumn{3}{c|}{\textbf{MNIST}} & \multicolumn{3}{c|}{\textbf{LGG}}   & \multicolumn{3}{c}{\textbf{WebKB}}          \\ 
\cmidrule(lr){2-10}
& ACC           & NMI           & PUR           & ACC           & NMI           & PUR    & ACC           & NMI           & PUR                                  \\
\midrule
DEMVC {\tiny\textcolor{gray}{[IS’21]}}   & 0.9734 &0.9587 &0.9734 & 0.5318 &0.2840 &0.5693 & 0.6156 &0.0158 &0.7812 \\
DSMVC {\tiny\textcolor{gray}{[CVPR’22]}}   & 0.9671 &0.9215 &0.9671 & 0.5993 &0.1721 &0.6105 & 0.7165 &0.2303 &0.7812 \\
DealMVC {\tiny\textcolor{gray}{[MM’23]}}   & 0.9804 &0.9593 &0.9804    & 0.6067 &\underline{0.4500} &0.6592 & 0.7755 &0.0170 &0.7812      \\
GCFAggMVC {\tiny\textcolor{gray}{[CVPR’23]}}  & 0.6949 &0.6768 &0.6949 & 0.5094 &0.0355 &0.5131 & 0.8259 &0.3031 &0.8259 \\
SCMVC {\tiny\textcolor{gray}{[TMM’24]}}   & 0.8610 &0.8263 &0.8610  & 0.5318 &0.1467 &0.5655 & 0.8230 &0.3702 &0.8230   \\
MVCAN {\tiny\textcolor{gray}{[CVPR’24]}}    & 0.9863 &0.9612 &0.9863  & \underline{0.6517} &0.4329 &\underline{0.6742} & 0.8202 &0.2023 &0.8202 \\
ACCMVC {\tiny\textcolor{gray}{[TNNLS’24]}}    & \underline{0.9886} &\underline{0.9659} &\underline{0.9886}  & 0.4981 &0.1611 &0.5618 & \underline{0.8696} &\underline{0.4863} &\underline{0.8696} \\
DMAC {\tiny\textcolor{gray}{[AAAI’25]}}    & 0.9720 &0.9302 &0.9720  & 0.4607 &0.0396 &0.5131 & 0.8145 &0.1844 &0.8145 \\
\midrule
MoEGCL (Ours) & \textbf{0.9920} & \textbf{0.9747} & \textbf{0.9920} &\textbf{0.7491} &\textbf{0.4534} &\textbf{0.7491}  & \textbf{0.9515} &\textbf{0.6503} &\textbf{0.9515}  \\
\bottomrule
\end{tabular}}
\label{tab:Clustering performance1}
\end{table*}

\begin{figure*}[htbp]
\centering
\subfigure[LGG] { 
\includegraphics[width=2.1in]{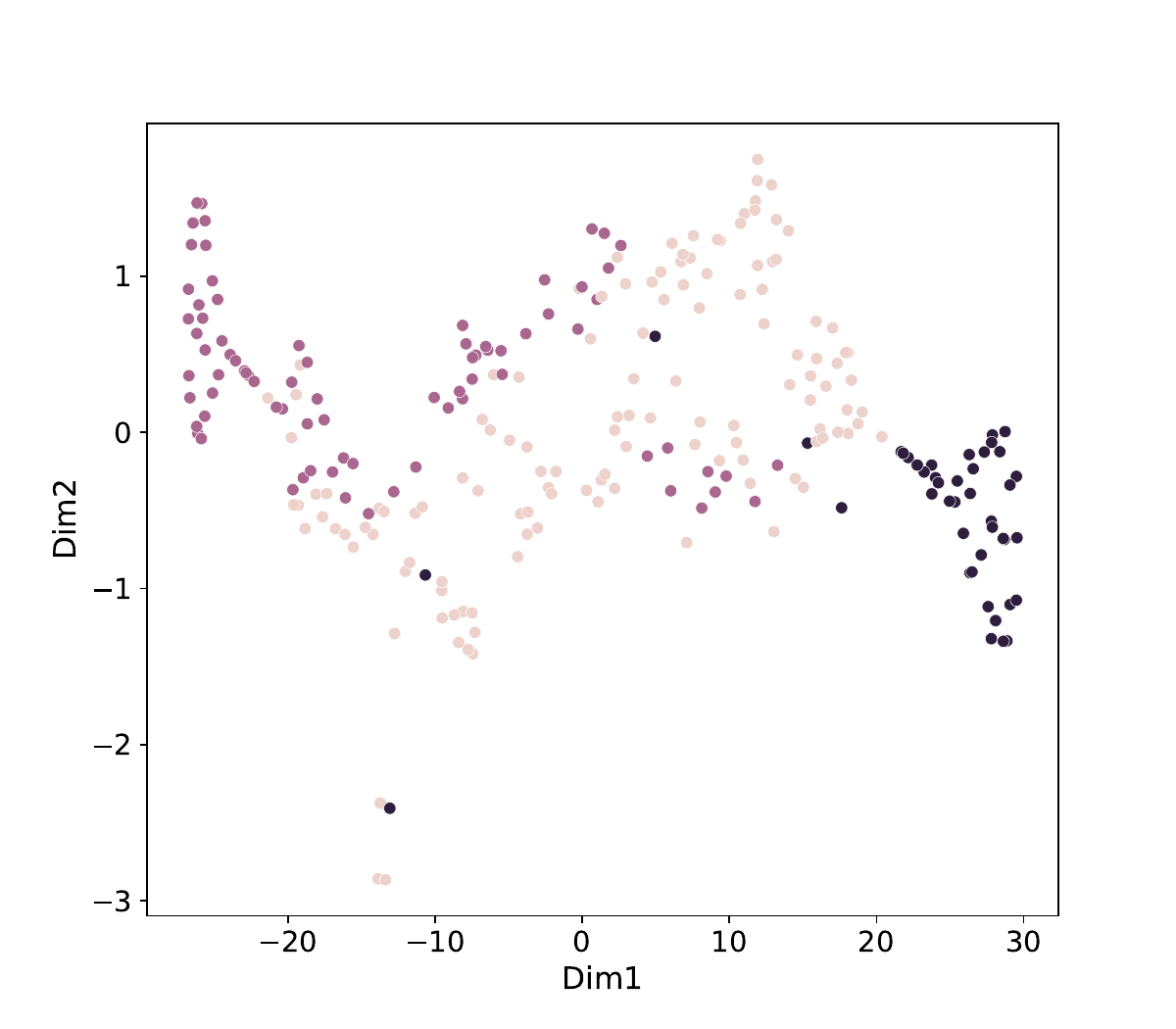}
}\hspace{0.24cm}
\subfigure[MNIST] { 
\includegraphics[width=2.1in]{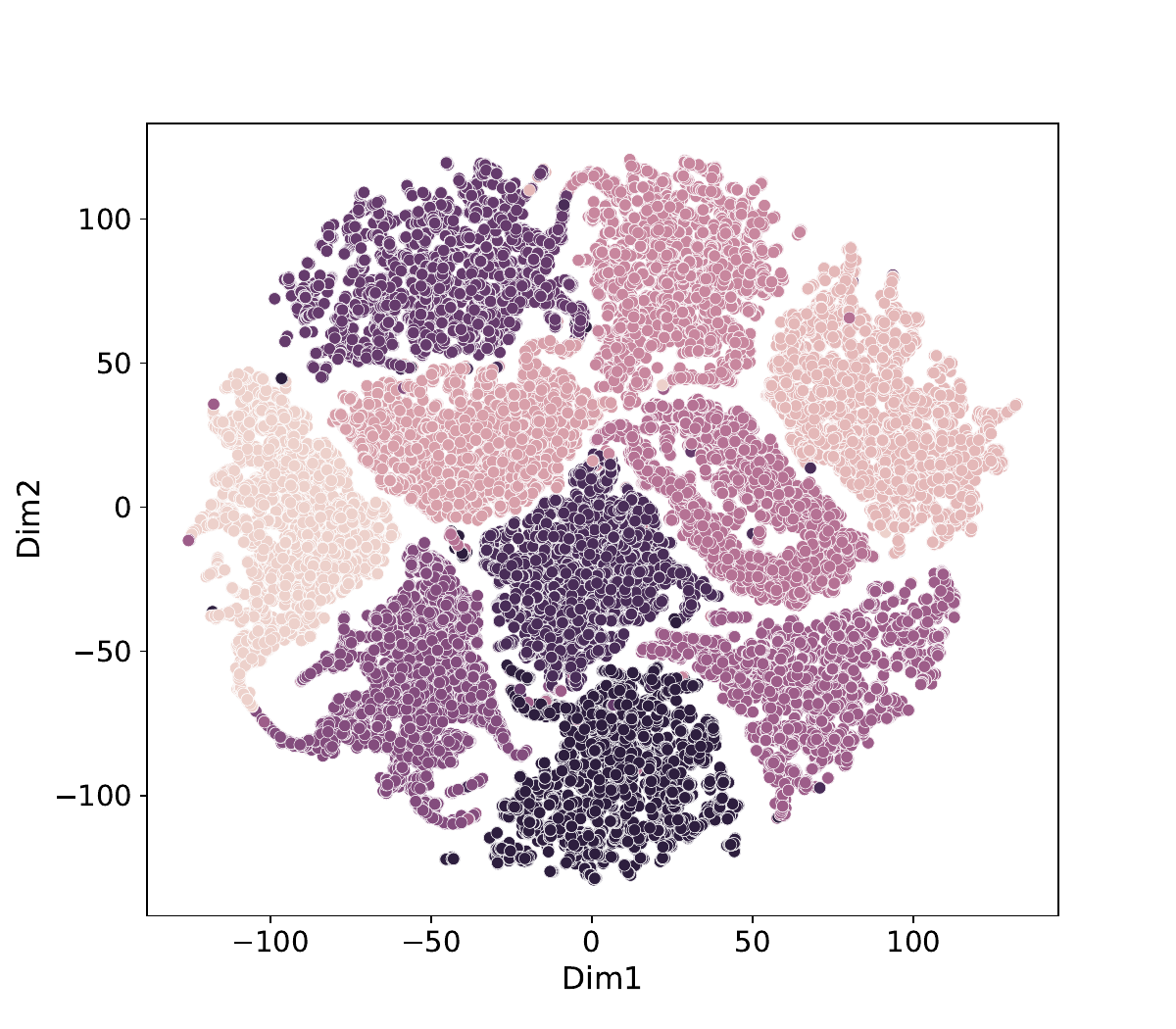}
}\hspace{0.2cm}
\subfigure[Caltech5V] { 
\includegraphics[width=2.1in]{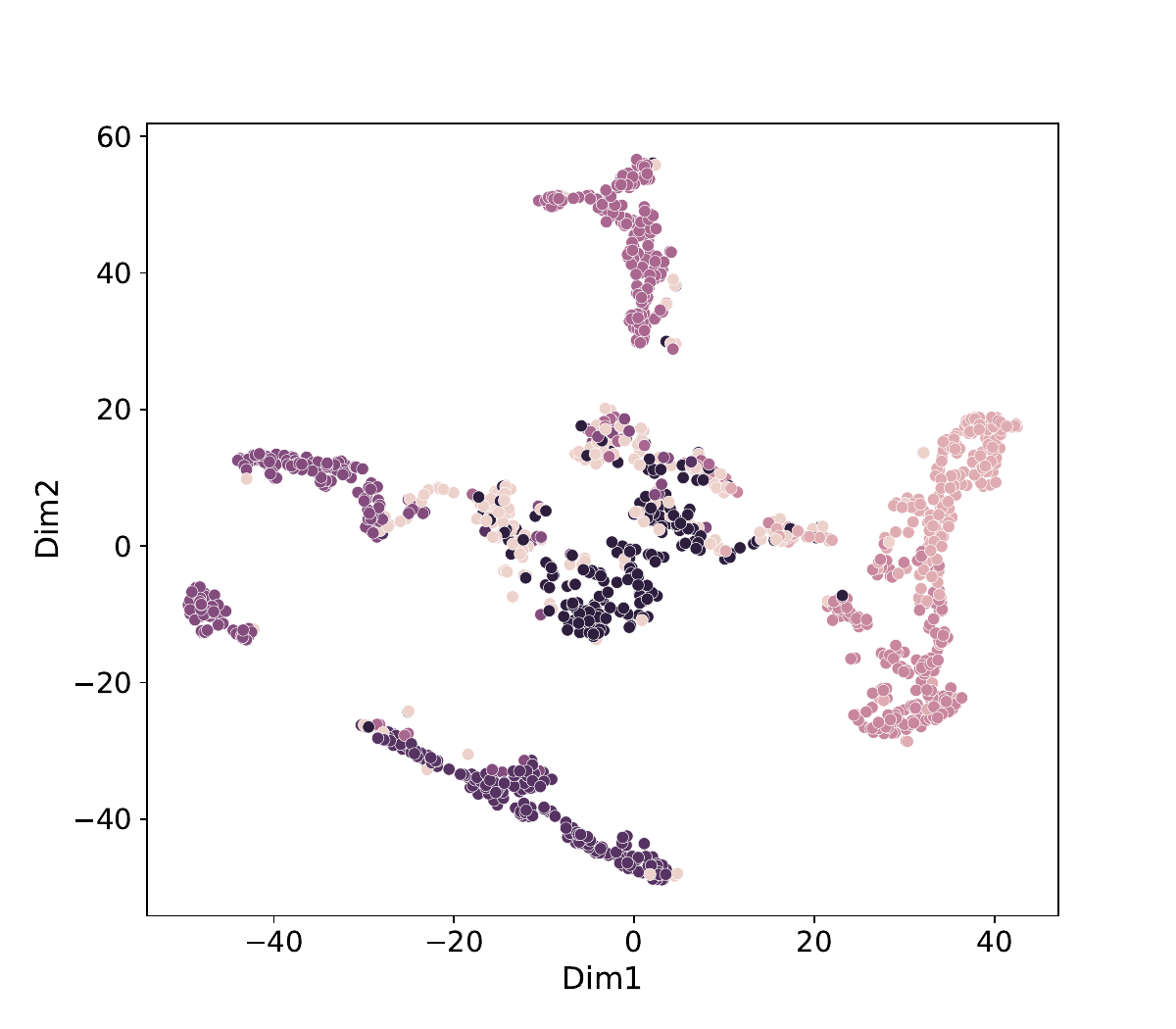}
}
\caption{The visualization results of the fused representations $\{\hat{h}_{i}\}^N_{i=1}$ on the LGG, MNIST, and Caltech5V datasets after convergence. From the visualization results of the three subfigures, it can be concluded that our proposed method separates the samples completely in the feature space, thus proving that our method is very effective.}
\label{fig:visualization} 
\end{figure*}
\begin{table*}[ht]
\renewcommand\arraystretch{1}
\small
\setlength{\tabcolsep}{9pt}
\setlength{\abovecaptionskip}{0.1cm}  %段前
\centering
\caption{Clustering result comparison on the Caltech5V, RGBD, and LandUse datasets. The best in bold, the second underlined.}
\resizebox{\textwidth}{!}{\begin{tabular}{l|ccc|ccc|ccc} 
\toprule
\multirow{2}{*}{\textbf{Methods}}   & \multicolumn{3}{c|}{\textbf{Caltech5V}}    & \multicolumn{3}{c|}{\textbf{RGBD}}               & \multicolumn{3}{c}{\textbf{LandUse}}          \\ 
\cmidrule(lr){2-10}
       & ACC           & NMI           & PUR   & ACC           & NMI           & PUR        & ACC           & NMI           & PUR   \\
\midrule
DEMVC {\tiny\textcolor{gray}{[IS’21]}} & 0.1693 &0.0443 &0.1786 & 0.3858 &0.3058 &0.4865 & 0.2057 &0.2032 &0.2171 \\
DSMVC {\tiny\textcolor{gray}{[CVPR’22]}}    & 0.5979 &0.4400 &0.6121 & 0.4341 &0.3740 &0.5418 & \underline{0.2695} &\underline{0.3494} &\underline{0.3048}    \\
DealMVC {\tiny\textcolor{gray}{[MM’23]}}   & 0.6157 &0.5081 &0.6157    & 0.4320 &0.2591 &0.4645  & 0.1895 &0.2014 &0.1929      \\
GCFAggMVC {\tiny\textcolor{gray}{[CVPR’23]}}  & 0.2893 &0.1264 &0.2979    & 0.3375 &0.3098 &0.4624     & 0.2610 &0.2991 &0.2833         \\
SCMVC {\tiny\textcolor{gray}{[TMM’24]}}   & 0.6086 &0.4210 &0.6100  & 0.3879 &0.3093 &0.5231 & 0.2595 &0.2791 &0.2752  \\
MVCAN {\tiny\textcolor{gray}{[CVPR’24]}} & \underline{0.7936} &\underline{0.6970} &\underline{0.7936}  & \underline{0.4458} &\underline{0.4134} &\underline{0.5977} & 0.2395 &0.3099 &0.2867   \\
ACCMVC {\tiny\textcolor{gray}{[TNNLS’24]}} & 0.6807 &0.5516 &0.7129  & 0.2926 &0.2695 &0.4500 & 0.2610 &0.2952 &0.2767   \\
DMAC {\tiny\textcolor{gray}{[AAAI’25]}} & 0.5850 &0.4609 &0.5907  & 0.3106 &0.2218 &0.4472 & 0.2429 &0.2899 &0.2786   \\
\midrule
MoEGCL (Ours) & \textbf{0.8207} & \textbf{0.7000} & \textbf{0.8207} &\textbf{0.4886} &\textbf{0.5516} &\textbf{0.6473}  & \textbf{0.3381} &\textbf{0.4206} &\textbf{0.3500}  \\
\bottomrule
\end{tabular}}
\label{tab:Clustering performance2}
\end{table*}

\begin{table}[t]
\small
\setlength{\tabcolsep}{7.5pt}
\setlength{\abovecaptionskip}{0.1cm}  %段前
\centering
\caption{Ablation study on the common MVC datasets.}
\begin{tabular}{ccccc} 
\toprule
Datasets                     & Method      & ACC & NMI  & PUR \\ 
\midrule
\multirow{5}{*}{Caltech5V} & w/o MoE & 0.7586 & 0.6241  & 0.7586 \\
                            & w/o GCN & 0.6857 & 0.5875 & 0.7164  \\
                            & w/o MoEGF & 0.4443 & 0.2934 & 0.4943  \\
                            & w/o EGCL & 0.5764 & 0.4757 & 0.5907 \\
                            & MoEGCL & \textbf{0.8207} & \textbf{0.7000} & \textbf{0.8207}  \\
\midrule
\multirow{5}{*}{RGBD}    & w/o MoE & 0.2202  & 0.1722& 0.3865 \\
                         & w/o GCN & 0.2961 & 0.2551 & 0.4741            \\
                         & w/o MoEGF & 0.4286  & 0.4071& 0.5845   \\
                         & w/o EGCL & 0.4845  & 0.5460& 0.6342   \\
                         & MoEGCL & \textbf{0.4886} & \textbf{0.5516} & \textbf{0.6473}  \\
\midrule
\multirow{5}{*}{MNIST} & w/o MoE & 0.9915 & 0.9735  & 0.9915 \\
                       &  w/o GCN & 0.9901 & 0.9704 & 0.9901 \\
                       & w/o MoEGF & 0.9817 & 0.9492 & 0.9817  \\
                       & w/o EGCL & 0.9905 & 0.9706 & 0.9905  \\
                        & MoEGCL       & \textbf{0.9920} & \textbf{0.9747} & \textbf{0.9920}  \\
\bottomrule
\end{tabular}
\label{tab:Ablation}
\end{table}
\subsection{Compared Methods} 
To evaluate the effectiveness of the proposed MoEGCL, we use eight recent state-of-the-art multi-view clustering methods as baselines. These methods are deep learning methods (including DEMVC \cite{xu2021deep}, DSMVC \cite{tang:58}, DealMVC \cite{yang:59}, GCFAggMVC \cite{Yan_2023_CVPR}, SCMVC \cite{wu:60}, MVCAN \cite{xu:61}, ACCMVC \cite{yan:72}, and DMAC \cite{wang:79}).

\begin{table}[t]
\small
\setlength{\tabcolsep}{7.5pt}
\setlength{\abovecaptionskip}{0.1cm}  %段前
\centering
\caption{Ablation study on the common MVC datasets.}
\begin{tabular}{ccccc} 
\toprule
Datasets                     & Method      & ACC & NMI  & PUR \\ 
\midrule
\multirow{5}{*}{WebKB} & w/o MoE & 0.7460 & 0.2247  & 0.7812 \\
                             &  w/o GCN & 0.5100 & 0.1112 & 0.7812 \\
                             & w/o MoEGF  & 0.5423 & 0.0435 & 0.7812  \\
                              & w/o EGCL  & 0.7888 & 0.2739 & 0.7888  \\
                             & MoEGCL & \textbf{0.9515} & \textbf{0.6503} & \textbf{0.9515}  \\
\midrule
\multirow{5}{*}{LGG} & w/o MoE & 0.5655 & 0.3094  & 0.6330 \\
                    &  w/o GCN & 0.5805 & 0.2017 & 0.5805 \\
                    & w/o MoEGF  & 0.4757 & 0.0968 & 0.5206  \\
                    & w/o EGCL  & 0.5880 & 0.1715 & 0.5880  \\
                    & MoEGCL    & \textbf{0.7491} & \textbf{0.4534} & \textbf{0.7491}  \\
\midrule
\multirow{5}{*}{LandUse} & w/o MoE & 0.3238 & 0.4038  & 0.3429 \\
                             & w/o GCN & 0.2652 & 0.3097 & 0.2767 \\
                             & w/o MoG  & 0.2514 & 0.3083 & 0.2876  \\
                             & w/o EGCL  & 0.3029 & 0.3912 & 0.3452  \\
                             & MoEGCL   & \textbf{0.3381} & \textbf{0.4206} & \textbf{0.3500}  \\
\bottomrule
\end{tabular}
\label{tab:Ablation2}
\end{table}

 \subsection{Implementation Details}
The source code of this paper is implemented using the PyTorch framework. To improve the generalization ability of the model, we set the dropout parameter of the network to $0.1$. During the model training process, we set the batch size 
$b$ to $256$. Model training is divided into two stages: pre-training and fine-tuning. The number of pre-training epochs $T_p$ is 200, and the number of fine-tuning epochs $T_f$ is $300$. The temperature coefficient $\tau$ for contrastive learning loss is set to 0.5. We unify the dimensions of the vectors outputted by the encoder and the input vectors for the contrastive loss, making $d_{\psi}$ equal to $512$ and $d_{\phi}$ equal to $128$. The learning rate of deep network training is $0.0003$. The combination coefficient ${\lambda}$ of the two loss functions is set to 1. The hardware platforms we used in our experiment are Intel(R) Xeon(R) Platinum 8358 CPU and Nvidia A40 GPU.

\begin{figure*}[htbp]
\centering
\subfigure{ 
\includegraphics[width=2.1in]{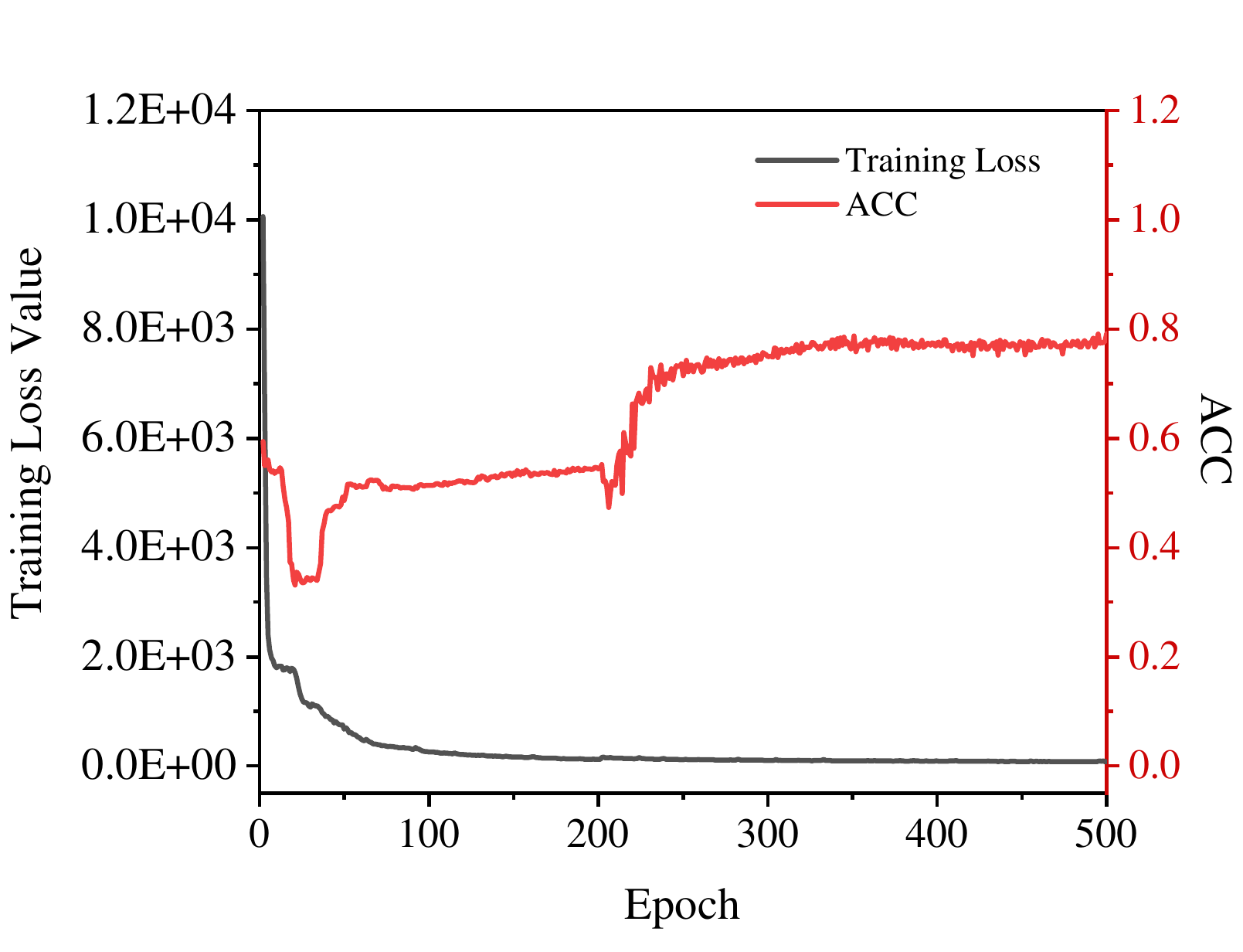}
}\hspace{0.1cm}
\subfigure{ 
\includegraphics[width=2.1in]{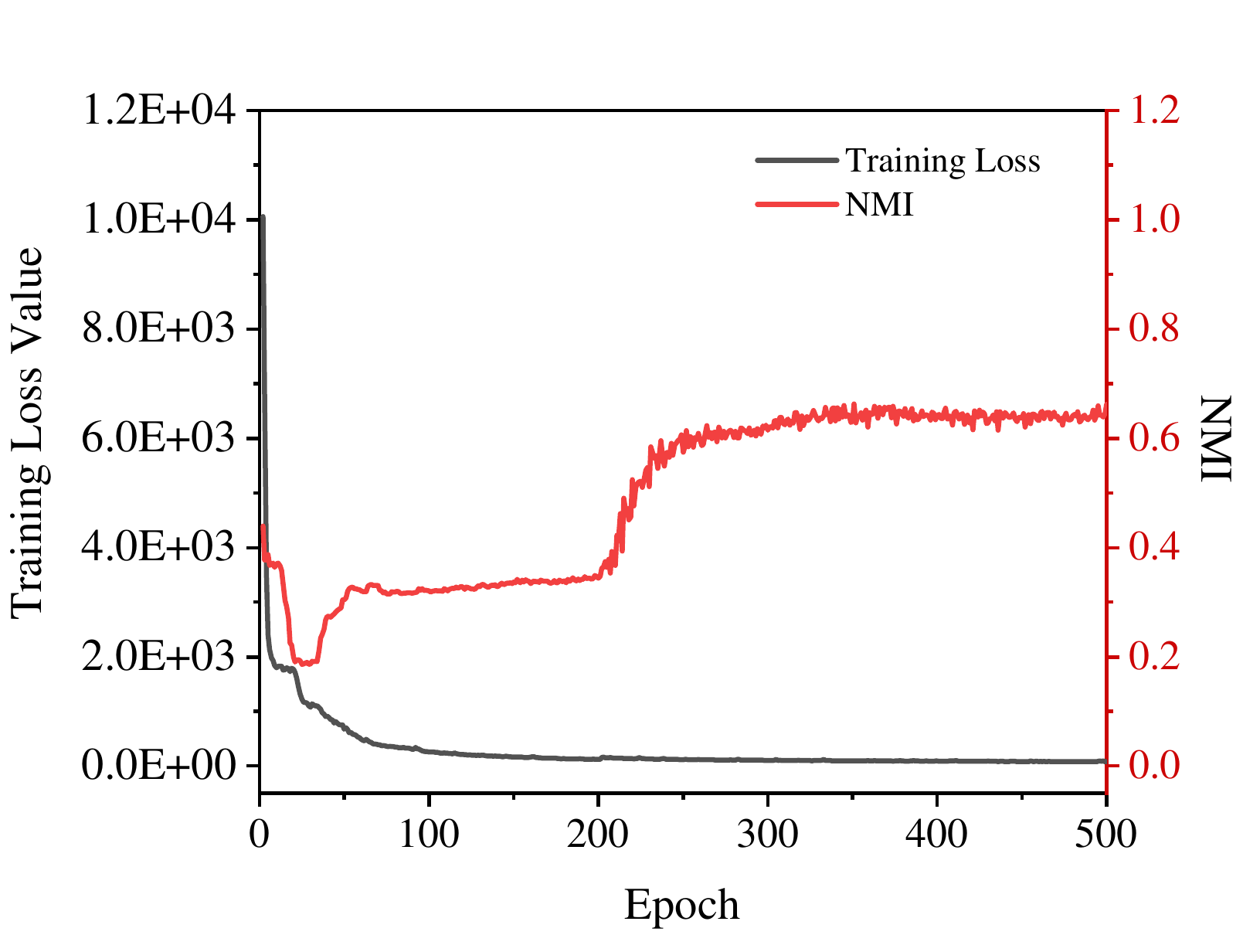}
}\hspace{0.1cm}
\subfigure{ 
\includegraphics[width=2.1in]{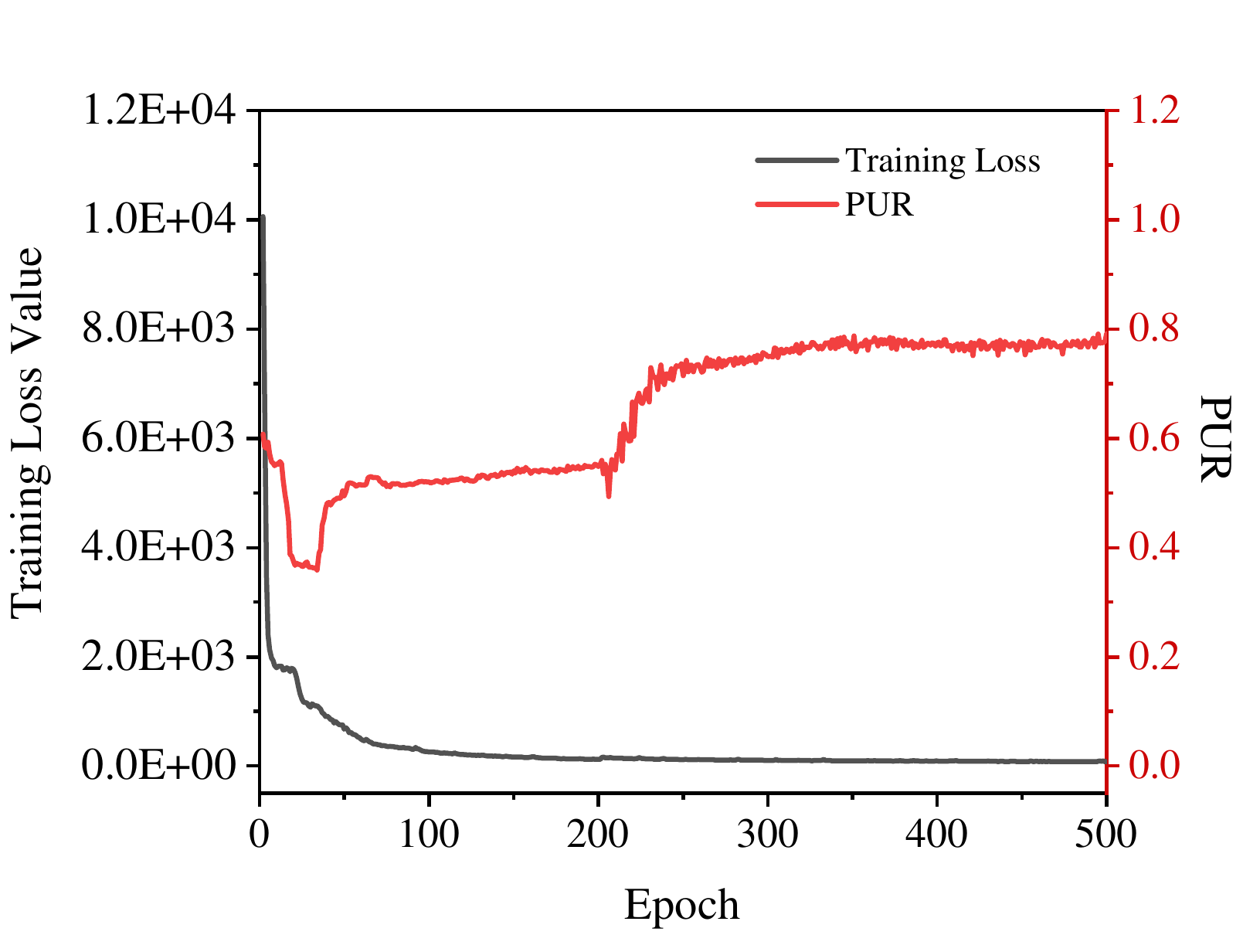}
}
\vspace{-1em}
\caption{The convergence analysis on the Caltech5V dataset. In the figure, the test ACC, NMI, and PUR are shown at the top, and the training loss is depicted at the bottom. It can be observed that around $400$ epochs, MoEGCL reaches a steady state, with the training loss no longer decreasing, and achieves the best accuracy compared with SOTA methods.}
\label{fig:convergence} 
\end{figure*}

\subsection{Evaluation Metrics}
We utilize three standard quantitative metrics, including unsupervised clustering accuracy (ACC), normalized mutual information (NMI), and purity (PUR). The larger these three indicators are, the better the clustering performance of the model.

\begin{figure*}[htbp]
\centering
\subfigure{ 
\includegraphics[width=1.95in]{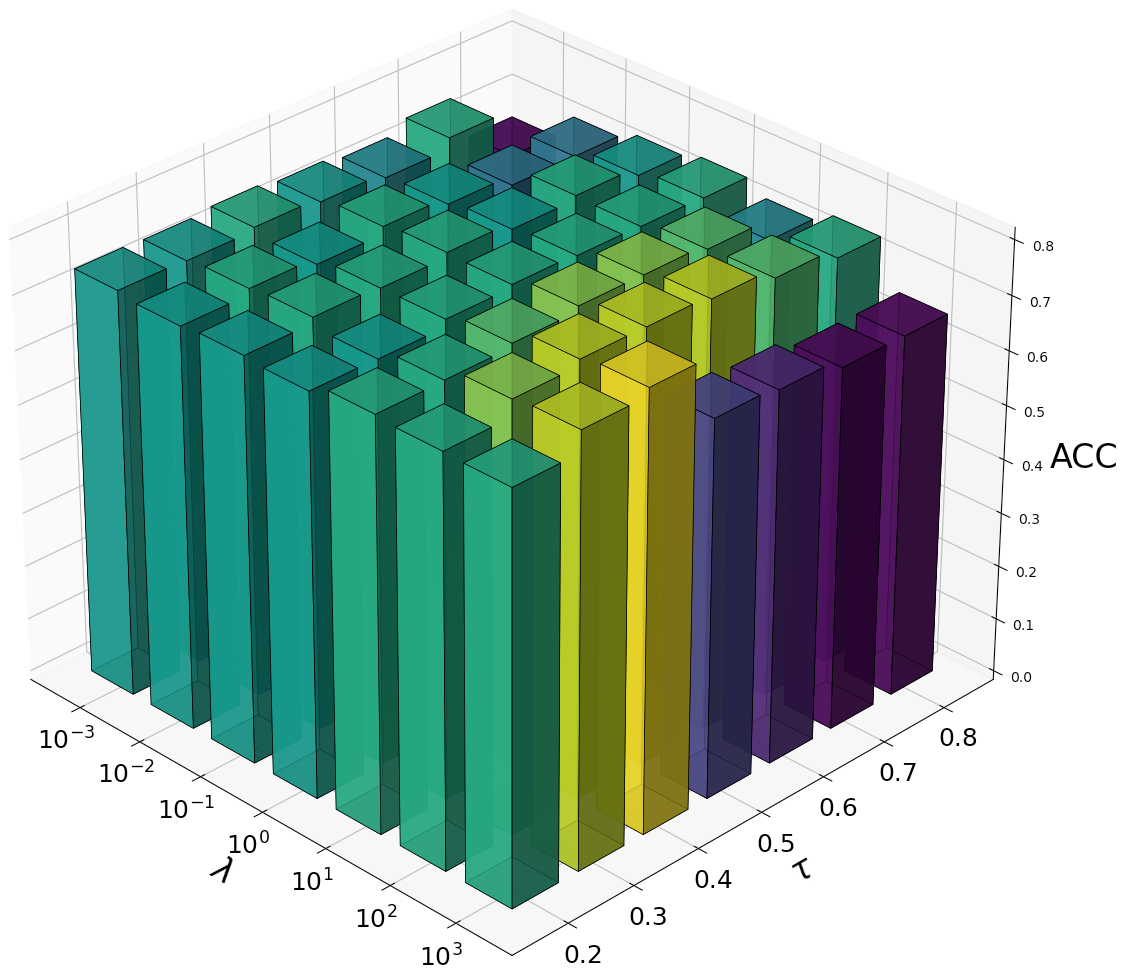}
}\hspace{0.4cm}
\subfigure{ 
\includegraphics[width=1.95in]{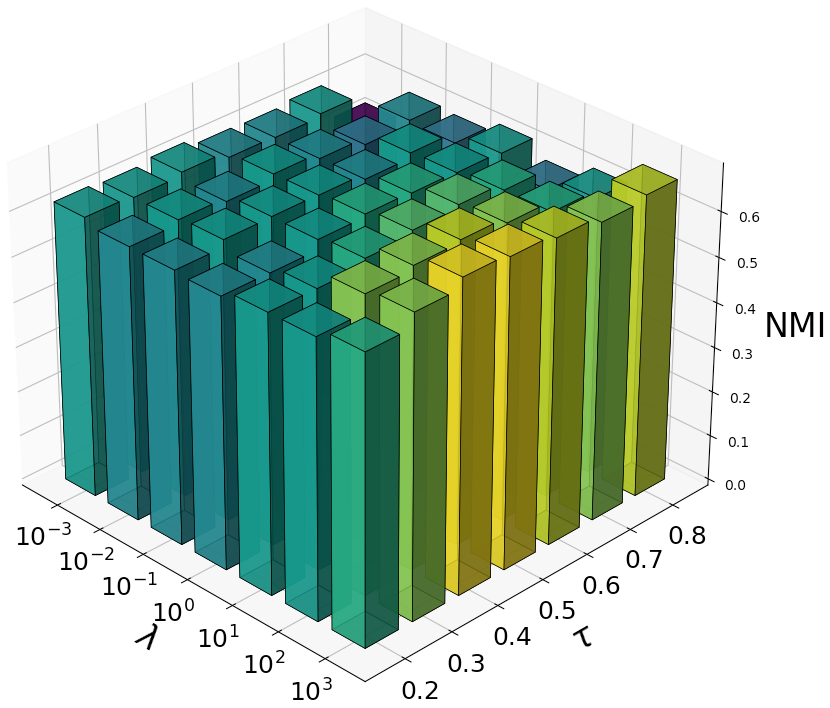}
}\hspace{0.4cm}
\subfigure{ 
\includegraphics[width=1.95in]{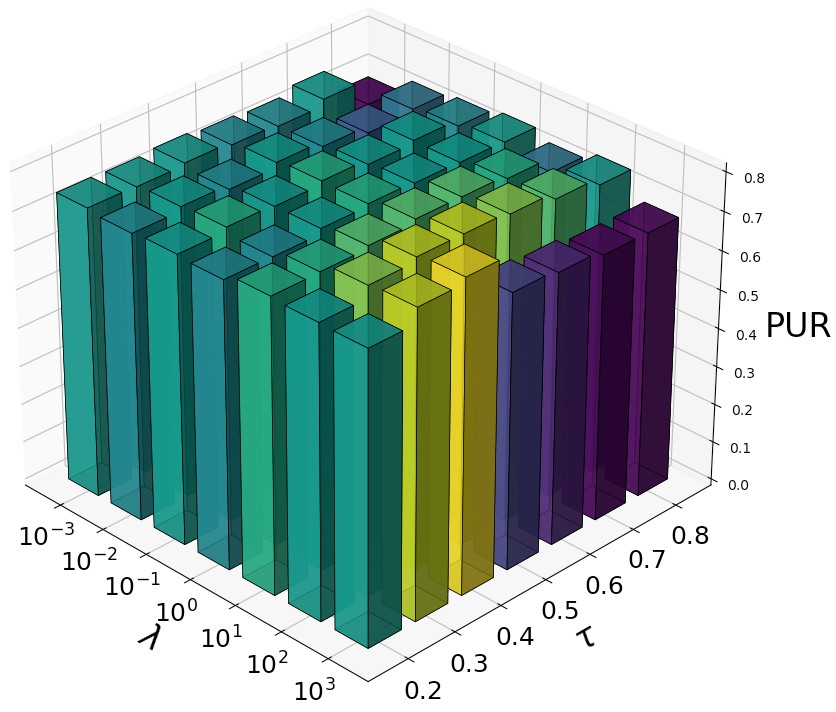}
}
\vspace{-1em}
\caption{The parameter analysis on the Caltech5V dataset. The figure shows the changes in three evaluation metrics: ACC, NMI, and PUR. The metrics are influenced by two hyperparameters $\lambda$ and $\tau$. $\lambda$ is the combination coefficient of two loss functions. $\tau$ denotes the temperature coefficient.}
\label{fig:hyper} 
\end{figure*}

\subsection{Experimental comparative results}
As shown in Table~\ref{tab:Clustering performance1} and Table~\ref{tab:Clustering performance2}, MoEGCL achieves superior performance compared to eight state-of-the-art baselines (DEMVC, DSMVC, DealMVC, GCFAggMVC, SCMVC, MVCAN, ACCMVC, and DMAC) across six benchmark datasets. In particular, we derive the following findings: On the WebKB dataset, MoEGCL surpasses the second-best method (ACCMVC) by $8.19\%$ in ACC. Similarly, on the RGBD dataset, our method outperforms MVCAN by $4.28\%$ in ACC, with consistent gains in NMI and PUR. In the other four datasets,  MoEGCL achieves state-of-the-art results on all remaining datasets, demonstrating robust generalizability. The performance advantage stems from two key innovations: 
\begin{itemize}
\item We present the MoEGF module in multi-view learning, which uses a Mixture-of-Experts network to achieve the fusion of ego graphs.
\item We propose the EGCL module to improve the representation similarity of samples in the cluster, which enhances fine-grained graph representation.
\end{itemize}

\subsection{Ablation Study}
\par
We conduct an ablation experiment to evaluate each component of the proposed MoEGCL on the six public datasets.

\textbf{Effectiveness of MoEGF module.} The ``w/o MoEGF" denotes removing the MoEGF module from the MoEGCL framework. The fused representation is set to $z_i$, which is the concatenation of all view-specific representations $\{z_i^m\}_{m=1}^M$. Table \ref{tab:Ablation} illustrates that, in the ACC term, the results of w/o MoEGF are $37.64$, $6.00$, $1.03$, $40.92$, $27.34$, and $8.67$ percent less than those of the MoEGCL. The results demonstrate that the MoEGF module significantly improves the performance of deep multi-view clustering.

\textbf{Validity of EGCL module.} The ``w/o EGCL" represents the elimination of the EGCL module in the MoEGCL framework. As illustrated in Table \ref{tab:Ablation}, the results of w/o EGCL are lower than those of the MoEGCL method by $24.43$, $0.41$, $0.15$, $16.27$, $16.11$, and $3.52$ percent on the ACC term. As opposed to merely using the same sample, the similarity of view presentation from the same cluster improves the fine-grained graph representation. Therefore, it improves the results in multi-view clustering tasks.

\textbf{Importance of MoE module.} In Table \ref{tab:Ablation}, the ``w/o MoE" indicates the removal of the MoE module and takes the average of the graph matrices for all views. We present the experimental results of ``w/o MoE" on six common datasets. On the WebKB dataset, compared with ``w/o MoE", MoEGCL achieves improvements of $20.55\%$, $42.56\%$, and $17.03\%$ on three evaluation metrics, respectively. Additionally, MoEGCL improves the three assessment metrics on the Caltech5V dataset by $6.21\%$, $7.59\%$, and $6.21\%$, respectively. Through the above experiments, it is proved that MoEGCL is superior to ``w/o MoE".

\textbf{Usefulness of GCN module.} As illustrated in Table \ref{tab:Ablation}, the ``w/o GCN" is the removal of the GCN module from the MoEGCL. It is shown that MoEGCL is better than ``w/o GCN" on all six public datasets.

\subsection{Visualization}
To further verify the effectiveness of MoEGCL,  we visualize the fused representations $\{\hat{h}_{i}\}^N_{i=1}$ using the t-SNE method \cite{van2008visualizing}, as shown in Figure \ref{fig:visualization}. We conduct visualization experiments on the three datasets: LGG, MNIST, and Caltech5V. 

In Figure \ref{fig:visualization}, $(a)$, $(b)$, and $(c)$ respectively represent the visualization results of the LGG, MNIST, and Caltech5V datasets. From the clustering results, it can be seen that the clusters after convergence are completely separated and the boundaries are relatively clear. There is basically no overlap between clusters. The visualization of Figure \ref{fig:visualization}(a) shows that the samples are divided into $3$ categories, and the visualization effect is very good, which can also correspond to the $3$ labels of the LGG dataset. Similarly, the visualization results of Figures \ref{fig:visualization}(b) and \ref{fig:visualization}(c) show $10$ categories and $7$ categories, which are consistent with the labels of the MNIST and Caltech5V datasets.

%Through the above visual analysis, the MoEGCL can perfectly separate sample data in the latent space, indicating that our method is very effective.

\subsection{Convergence Analysis and Generalization Ability} 
The convergence and generalization capabilities of MVC are crucial. To validate the convergence and generalization effectiveness of MoEGCL, we conduct tests on the Caltech5V dataset. The performance and loss are shown in Figure \ref{fig:convergence}.

Figure \ref{fig:convergence} consists of three subfigures, which illustrate the variation curves of test ACC, NMI, PUR, and training loss. The loss curve steadily decreases as training progresses, stabilizing at $300$ epochs. This indicates that the deep multi-view clustering network is functioning effectively. Right from the start of the experiment, the test ACC,  NMI, and PUR for the evaluation measure show a significant increase. After $400$ epochs, the ACC, NMI, and PUR stabilize and remain unchanged with further training, indicating robust generalization capabilities and the absence of overfitting. Similar convergence and generalization results are observed across multiple datasets.

%Through the experimental analysis presented above, we demonstrate that MoEGCL exhibits strong convergence and generalization ability.

\subsection{Parameter Analysis}
As shown in Figure \ref{fig:hyper}, we conduct experiments on hyperparameter analysis. Specifically, we analyze the impact of two hyperparameters, $\lambda$ and $\tau$, on clustering evaluation metrics. $\lambda$ represents the combination coefficient of two loss functions. $\tau$ denotes temperature coefficient. The evaluation metrics is the test ACC, NMI, and PUR. We conduct hyperparameter experiments on the Caltech5V dataset. The value of $\lambda$ varies from $10^{-3}$ to $10^{3}$. The values of $\tau$ are $0.2$, $0.3$, $0.4$, $0.5$, $0.6$, $0.7$, and $0.8$. We draw three subfigures representing the changes in the three metrics ACC, NMI, and PUR under the influence of hyperparameters. Figure \ref{fig:hyper} shows that ACC, NMI, and PUR are not significantly affected by the two hyperparameters and exhibit relatively stable performance.

%Through the experiments of hyperparameter analysis mentioned above, we conclude that the proposed MoEGCL is insensitive to parameters, thus proving that it is a robust method.

\section{Conclusion and Future Work}
This paper proposes a novel Mixture of Ego-Graphs Contrastive Representation Learning (MoEGCL) framework. It aims to implement fine-grained graph fusion for deep multi-view clustering tasks. It is composed of two modules, Mixture of Ego-Graphs Fusion (MoEGF) and Ego Graph Contrastive Learning (EGCL). The MoEGF module, which constructs ego graphs and utilizes the Mixture-of-Experts network, is presented for the fusion of ego graphs at the sample level. Furthermore, the EGCL module is proposed to lessen conflicts in the same clusters and get better semantic structure information. It further improves fine-grained graph representation. Extensive experiments demonstrate that MoEGCL achieves superior performance over state-of-the-art methods in deep multi-view clustering tasks. Specifically, we compare eight baseline methods on six public datasets to verify the effectiveness of our proposed method. In summary, MoEGCL is a simple and effective representation learning method, it has great application prospects.

\newpage
\section*{Acknowledgments}
This work is supported by the National Key Research and Development Program of China (Grant No. 2021ZD0201501), the National Natural Science Foundation of China (No. 22574146, No. 32200860, and No. 62306289), by the Regional Innovation and Development Joint Fund of National Natural Science Foundation of China (U22A6001), the National Key R\&D Program of China (2022YFB4500405), the Youth Foundation Project of Zhejiang Province (Grant No. LQ22F020035), the ``Pioneer'' and ``Leading Goose'' R\&D Program of Zhejiang (2024SSYS0007), the National Key R\&D Program of China (Grant No. 2024YFB4505602), by Anhui Provincial Natural Science Foundation (Grant No.2408085QF214), the Fundamental Research Funds for the Central Universities (Grant No.WK2100000045, No.WK2080000206), and the Opening Project of the State Key Laboratory of General Artificial Intelligence (Grant No.SKLAGI2024OP10, Grant No.SKLAGI2024OP11).

\bibliography{aaai2026}

\end{document}